\documentclass[10pt,twocolumn,letterpaper]{article}

\usepackage[pagenumbers]{iccv} % To force page numbers, e.g. for an arXiv version
\usepackage{url}
\usepackage{graphicx}
\usepackage{paralist}
\usepackage{bm}
\usepackage{amsmath}
\usepackage{multirow}
\usepackage{xr}
\usepackage{booktabs}
\usepackage[leftcaption]{sidecap}
\usepackage{booktabs} % For formal tables
\usepackage{sidecap}
\usepackage{subcaption}
\usepackage{stfloats}
\usepackage{float}
\usepackage{xcolor}

\definecolor{iccvblue}{rgb}{0.21,0.49,0.74}
\usepackage[pagebackref,breaklinks,colorlinks,allcolors=iccvblue]{hyperref}

\def\paperID{1040} % *** Enter the Paper ID here
\def\confName{ICCV}
\def\confYear{2025}

\title{DreamDance: Animating Character Art via Inpainting Stable Gaussian Worlds}

\author{
    Jiaxu Zhang$^{1, 2, 3}$\qquad
    Xianfang Zeng$^{3}$\footnotemark[2]\qquad
    Xin Chen$^{4}$\qquad \\
    Wei Zuo$^{3}$\qquad 
    Gang Yu$^{3}$\footnotemark[3] \qquad
    Guosheng Lin$^{2}$ \qquad
    Zhigang Tu$^{1}$\footnotemark[3]\\
    $^{1}$Wuhan University\qquad
    $^{2}$Nanyang Technological University \qquad
    $^{3}$StepFun \qquad
    $^{4}$ByteDance \qquad \\
    {\tt \small Project page: \url{https://kebii.github.io/DreamDance}}
    }

\begin{document}

% \maketitle

\twocolumn[{%
    \centering
    \vspace{-1.4cm}
    \includegraphics[width=1.8cm]{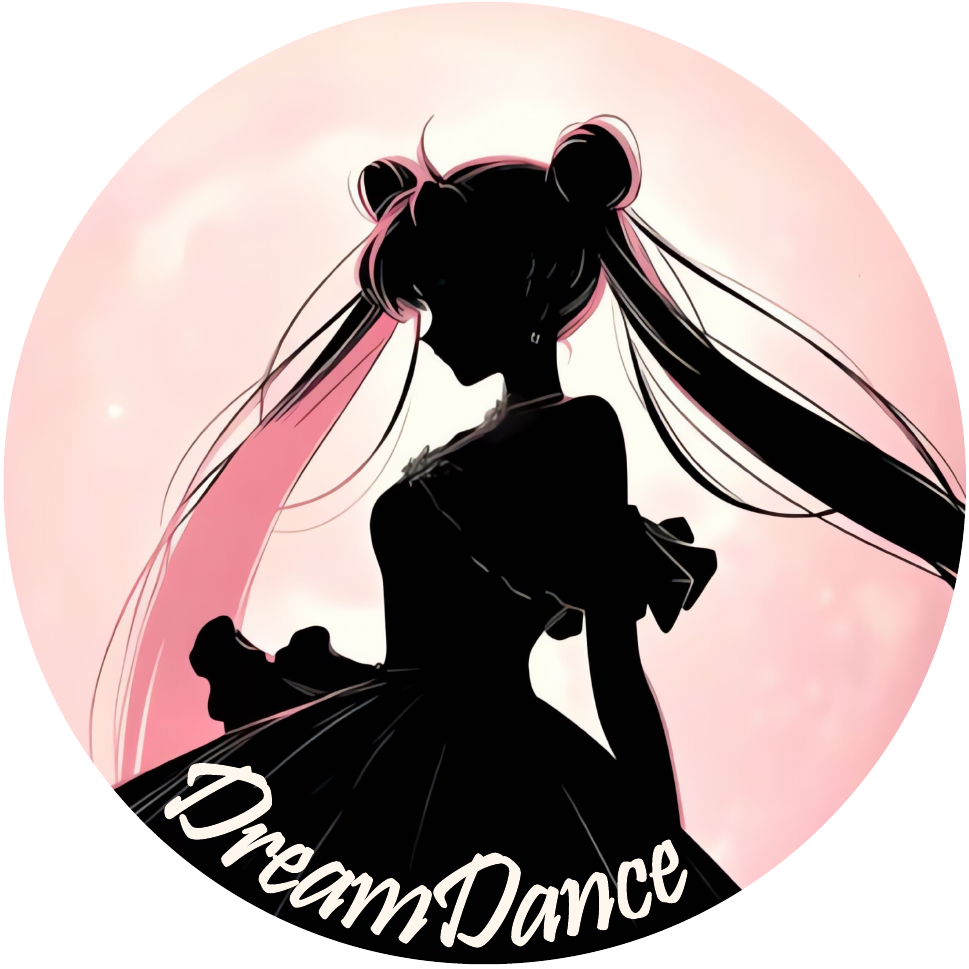}\\[\baselineskip] % 调整宽度控制logo大小
    \vspace{-1.5cm} % 控制logo和标题之间的间距
    \maketitle
}]

\renewcommand{\thefootnote}{\fnsymbol{footnote}}
% \footnotetext[1]{Most of this work was done during Jiaxu’s internship at StepFun.}
\footnotetext[2]{Xianfang Zeng is the project leader.}
\footnotetext[3]{Corresponding authors: skicy@outlook.com, tuzhigang@whu.edu.cn}
\renewcommand{\thefootnote}{\arabic{footnote}}

\begin{abstract}
This paper presents DreamDance, a novel character art animation framework capable of producing stable, consistent character and scene motion conditioned on precise camera trajectories.
To achieve this, we re-formulate the animation task as two inpainting-based steps: Camera-aware Scene Inpainting and Pose-aware Video Inpainting.
The first step leverages a pre-trained image inpainting model to generate multi-view scene images from the reference art and optimizes a stable large-scale Gaussian field, which enables coarse background video rendering with camera trajectories.
However, the rendered video is rough and only conveys scene motion. To resolve this, the second step trains a pose-aware video inpainting model that injects the dynamic character into the scene video while enhancing background quality.
Specifically, this model is a DiT-based video generation model with a gating strategy that adaptively integrates the character's appearance and pose information into the base background video.
Through extensive experiments, we demonstrate the effectiveness and generalizability of DreamDance, producing high-quality and consistent character animations with remarkable camera dynamics.

\end{abstract}    
\section{Introduction}
\begin{figure}[t]
% \vspace{-.4cm}
\setlength{\abovecaptionskip}{-.2cm}
\setlength{\belowcaptionskip}{-.3cm}
\begin{center}
   \includegraphics[width=1.0\linewidth]{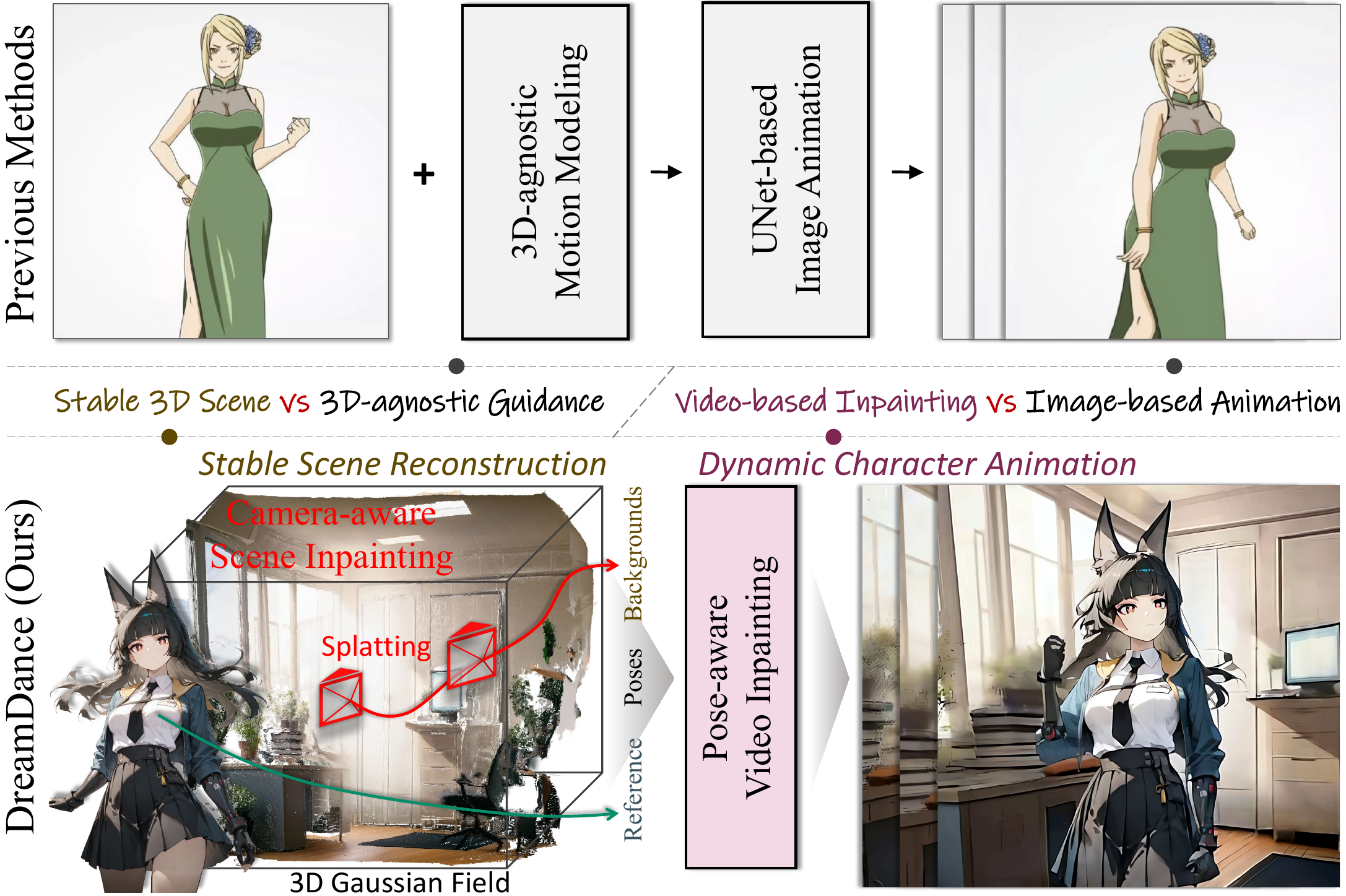}
\end{center}
   \caption{\textbf{We propose DreamDance}, a novel paradigm that re-formulates the character art animation task into two inpainting-based steps: \textit{Camera-aware Scene Inpainting} for stable scene reconstruction and \textit{Pose-aware Video Inpainting} for dynamic character animation.}
\label{fig:1}
\end{figure}

Animating character art is a fundamental challenge in the 2D animation industry, with a wide range of applications in film, game, and digital design. However, traditional 2D animation is a labor-intensive and time-consuming process that requires expertise in professional software such as MMD \cite{MikuMikuDance} and Live2D \cite{Live2D}. Recently, human video generation methods, particularly MikuDance \cite{zhang2024mikudance}, have revolutionized this challenging task, making it accessible to non-experts. Derived from previous methods \cite{bhunia2023person, zhu2024champ, hu2024animate}, MikuDance performs two strategies to animate character art using driving videos, as illustrated in Figure \ref{fig:1}. 

The first strategy is motion modeling, which uses pose image sequences to drive characters and 2D scene flow to guide backgrounds. Similar to MikuDance, existing methods incorporate other motion guidance, such as optical flow \cite{wang2024leo, niu2024mofa} and camera parameters \cite{shao2024human4dit, wang2024humanvid}, to represent global camera movements. However, this motion guidance is entirely 3D-agnostic and struggles to provide consistent scene context, leading to scene distortion during large-scale camera movements. This inconsistency arises from the implicit inpainting process, where the scene dynamics exceed the area covered by the reference, requiring the model to hallucinate missing regions. Therefore, it is crucial to explore 3D-aware scene modeling for consistent camera control.

The second strategy is image animation, which utilizes the UNet-based Stable Diffusion model \cite{ronneberger2015u, rombach2022high} to animate the reference character art using mixed motion guidance. Additionally, some recent methods have developed incremental pose encoders \cite{zhu2024champ, zhang2024mimic} and reference adapters \cite{wang2024disco, chang2025x}, using SVD \cite{blattmann2023stable} as a base model to achieve human image animation. However, due to the limitations of base model capacity and the ambiguity of mixed guidance, the animation results from these methods exhibit significant temporal jitters in both characters and backgrounds. Therefore, it is crucial to introduce a more powerful video foundation model and redefine the animation process with explicit contextual scene guidance.

Unlike MikuDance and other relevant methods, we propose DreamDance, a new paradigm for animating in-the-wild character art. As illustrated in Figure \ref{fig:1}, DreamDance reformulates the motion modeling and image animation processes into two inpainting-based steps: Camera-aware Scene Inpainting and Pose-aware Video Inpainting. These two components work synergistically to generate consistent, high-quality animation sequences from the reference character art and driving videos.

Camera-aware Scene Inpainting is presented for stable scene reconstruction. 
Inspired by existing 3D Gaussian methods \cite{yu2024wonderworld, chung2023luciddreamer, wang2024vistadream}, we leverage a pre-trained image inpainting model \cite{Fooocus} to generate multi-view images and construct a large-scale 3D Gaussian field from the reference character art. This process utilizes both a pre-defined spiral camera trajectory and an extracted custom camera trajectory. A coarse background video is then rendered along the custom trajectory by splatting the stable Gaussian field. This background video contains consistent scene motion information and serves as a rough yet foundational video for the later character animation stage.
% This background video, resembling a pose-based inpainting mask, forms the foundational video basis for the next character animation stage.

Pose-aware Video Inpainting is proposed for injecting dynamic character animation into the coarse scene video.
Based on a video generation model, i.e., CogVideoX \cite{yang2024cogvideox}, we train a gated video inpainting model to refine the coarse backgrounds and inject the reference character according to the pose guidance. The gating strategy is designed to adaptively incorporate both the character's appearance and poses based on the denoising time step, ensuring character and background consistency throughout the animation. Additionally, we exploit a 3D Gaussian-free training approach to train the dynamic video inpainting model directly using background-degraded video datasets.

By leveraging this new paradigm, DreamDance animates diverse character art with stable scenes and precise camera movements, generating spatio-temporally consistent animations. We evaluate DreamDance using a wide range of reference character art and driving videos. Both qualitative and quantitative results demonstrate that DreamDance can generate high-quality animations, particularly with flexible and coherent scene dynamics.

Contributions of DreamDance are listed in three folds:
\begin{compactitem}
\item Camera-aware Scene Inpainting and Pose-aware Video Inpainting are proposed to re-formulate the character art animation task, enabling explicit and consistent scene context modeling.
\item A gating strategy is introduced into a fundamental video generation model to achieve adaptive video inpainting, enabling high-dynamic animation of character art within a stable Gaussian scene.
\item Extensive experiments demonstrate the effectiveness and generalizability of DreamDance, achieving superior animation quality over state-of-the-art methods.
\end{compactitem}

\section{Related Work}
\noindent\textbf{2D character animation} provides a vibrant platform for storytelling but has long been a challenge in the animation industry. Some previous methods construct animated 3D characters from reference images and re-project them into 2D videos \cite{smith2023method, zhou2024drawingspinup, zhangtapmo, qiu2024anigs}. These methods require precise geometry, rigging, and motion editing, making them hard to automate, and often resulting in a loss of the original 2D style. Recent approaches like Textoon \cite{he2025textoon} and AniClipart \cite{wu2024aniclipart} aim to generate animatable 2D characters using image and video generation models. However, they still require significant manual work, and the character’s motion freedom is limited. In contrast, MikuDance \cite{zhang2024mikudance} directly generates 2D animation through an image animation model, offering a promising solution, but it faces issues with scene distortion and artifacts due to its 3D-agnostic approach. Derived from MikuDance, we propose DreamDance, which introduces two inpainting steps for 3D context-aware and consistent character art animation in stable Gaussian scenes.

\vspace{1mm}
\noindent\textbf{Human image animation} has gained popularity in recent years \cite{bhunia2023person, lin2025omnihuman}, with many methods building on pre-trained image and video generation models \cite{rombach2022high, blattmann2023stable}. For example, Animate Anyone \cite{hu2024animate} uses a reference-denoising UNet structure and a temporal module from AnimateDiff \cite{guoanimatediff} to improve video consistency. DisCo \cite{wang2024disco} separates human subjects from backgrounds, allowing for more flexible combinations. MimicMotion \cite{zhang2024mimic} introduces a regional loss based on pose confidence to enhance human fidelity. Animate-X \cite{tan2024animate} extends the pipeline to anthropomorphic characters. However, these methods mainly focus on character actions and overlook scene dynamics.

Moving forward, Human4DiT \cite{shao2024human4dit} and HumanVid \cite{wang2024humanvid} address camera movements by incorporating a camera encoder. However, camera guidance alone cannot handle large-scale scene dynamics, as it is difficult for the model to consistently fill in missing areas. MIMO \cite{men2024mimo} and Animate Anyone 2 \cite{hu2025animate2} capture environmental representations from the driving video and restore the backgrounds in the animation, but they focus on different goals than 2D character animation, where the scene comes from the reference character art. In this work, we reconstruct 3D Gaussian scenes by inpainting reference character art to support precise camera dynamics, and use an MM-DiT-based video foundation model to generate high-quality character animations.

\vspace{1mm}
\noindent\textbf{Video inpainting} typically focuses on two main tasks: object removal and text-guided inpainting. Traditional methods, like E$^2$FGVI \cite{li2022towards} and FGT \cite{zhang2022flow}, use optical flow-guided feature propagation to reconstruct missing areas with coherent content. More recently, models like AVID \cite{zhang2024avid} and CoCoCo \cite{zi2024cococo} have integrated pre-trained generative inpainting models \cite{rombach2022high} with motion modules for text-guided video inpainting. Unlike these approaches, our video inpainting step focuses on filling in the coarse background with animated characters guided by pose videos. To achieve this, we propose a gating strategy within the DiT \cite{peebles2023scalable, yang2024cogvideox} model to adaptively integrate the character's appearance and poses, enabling dynamic character animation.

\vspace{1mm}
\noindent\textbf{3D Gaussian Splatting} \cite{kerbl20233d} utilizes the concept of Gaussian splats combined with spherical harmonics and opacity to represent 3D scenes. Later work incorporates image inpainting to generate multi-view images and reconstruct 3D Gaussian fields from a single image \cite{chung2023luciddreamer, wang2024vistadream, yu2024wonderworld, liang2024wonderland, zou2024triplane}. Inspired by these approaches, we reconstruct stable Gaussian scenes from the reference character art and render coarse background videos to improve scene consistency.
\begin{figure*}[t]
\vspace{-.6cm}
\setlength{\abovecaptionskip}{-.2cm}
\setlength{\belowcaptionskip}{-.4cm}
\begin{center}
   \includegraphics[width=1.0\linewidth]{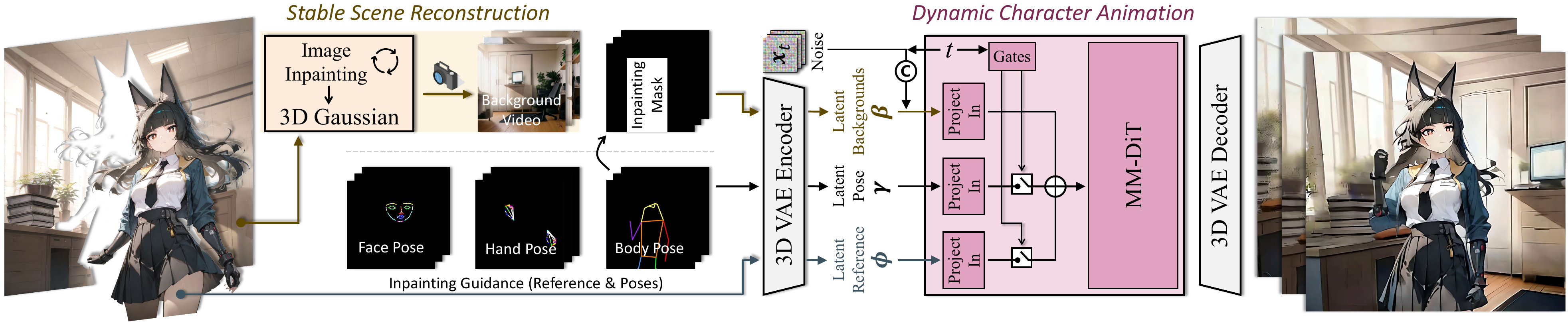}
\end{center}
   \caption{\textbf{Illustration of our DreamDance.} The reference character art is decomposed into foreground and background layers. The background image is used to reconstruct a stable 3D Gaussian scene through a wrap-and-inpaint scheme, enabling coarse background video rendering based on custom camera trajectories. The gated MM-DiT model then inpaints the background video based on the foreground character and the driving poses, generating dynamic character animations. }
\label{fig:2}
\end{figure*}

\section{Method}
As illustrated in Figure \ref{fig:2}, given a character art $\mathcal{I}$ and a driving video $\mathcal{V}$, the goal of DreamDance is to animate the image $\mathcal{I}$ based on the human and camera motion in the video $\mathcal{V}$. Specifically, we utilize Xpose \cite{yang2023Xpose} to separately extract the pose sequences of the human body, face, and hand, and employ DPVO \cite{teed2024deep} to extract the camera poses $\{\bm{p}^{c}_l\}^L_{l=1}$, $\bm{p}^{c}\in \mathbb{R}^{L \times 7}$ from $\mathcal{V}$. $L$ indicates the sequence length. The character and the background are segmented from $\mathcal{I}$ using BiRefNet \cite{zheng2024bilateral}. Next, we reconstruct a 3D Gaussian field from the reference background through multi-view image inpainting, using both a pre-defined spiral camera trajectory and the extracted camera trajectory. Then, a coarse background video is rendered by splitting the 3D Gaussian field according to the extracted camera poses, and an inpainting mask is generated from the driving pose sequence. Finally, all the references and guidance are processed using the pre-trained VAE and input into the gated video inpainting model for dynamic character animation.

\subsection{Preliminaries}

\noindent\textbf{Diffusion Denoising Probabilistic Models.} Diffusion-based generative models \cite{sohl2015deep, ho2020denoising} represent the data distribution by constructing a Markov chain. Given an input data distribution $\bm{x}_{0}$, the forward process applies a Markov noising process of $T$ steps on $\bm{x}_{0}$ to obtain $\{\bm{x}_{t}\}^{T}_{t=0}$:
\begin{equation} \label{preEq.1}
q(\bm{x}_{t}|\bm{x}_{t-1}) = \mathcal{N} \left (\sqrt{\alpha_{t}}\bm{x}_{t-1}, (1-\alpha_{t})\mathbf{I} \right ),
\end{equation}
where $\alpha_{t} \in (0,1)$ are constant hyper-parameters. When $\alpha_{t}$ is small enough, $\bm{x}_{T} \sim \mathcal{N}(\mathbf{0},\mathbf{I})$.
The reverse process takes a noisier data distribution $\bm{x}_{t}$ and generates a less noisy distribution $\bm{x}_{t-1}$ using a noise predictor, which is trained with the simple loss function:
\begin{equation} \label{preEq.2}
\mathcal{L}_{simple} := \mathbb{E}_{\bm{\epsilon}, t, c} \left [\lVert \bm{\epsilon}- \bm{\epsilon}_{\theta}(\bm{x}_{t},t,c)\rVert^{2}_{2} \right ],
\end{equation}
where $\bm{\epsilon}$ is the Gaussian noise. $c$ is the text condition. $\bm{\epsilon}_{\theta}(\cdot)$ is the trainable noise predictor. In this work, we utilize the pre-trained CogVideoX, an MM-DiT-based video diffusion model, as the base model to achieve pose-aware video inpainting in DreamDance.

\vspace{1mm}
\noindent\textbf{3D Gaussain Splatting (3DGS).} Prior works \cite{kerbl20233d, zwicker2001ewa} propose to represent a 3D scene as a set of scaled 3D Gaussian primitives $\{\mathcal{G}_{k}|k=1,...,K\}$ and render scene images using volume splitting. The geometry of each scaled 3D Gaussian $\mathcal{G}_{k}$ is parameterized by an opacity $\alpha_{k} \in [0, 1]$, center $\bm{o}_{k} \in \mathbb{R}^{3 \times 1}$, and covariance matrix $\bm{\Sigma}_{k} \in \mathbb{R}^{3 \times 3}$.

To render an image for a given camera defined by rotation $\mathbf{R} \in \mathbb{R}^{3 \times 3}$ and translation $\mathbf{t} \in \mathbb{R}^{3}$, the 3D Gaussians are first transformed into camera coordinates:
\begin{equation} \label{preEq.3}
\bm{o}_{k}^{'} = \mathbf{R}\bm{p}_{k} + \mathbf{t}, \quad \bm{\Sigma}_{k}^{'} = \mathbf{R}\bm{\Sigma}_{k}\mathbf{R}^{T}.
\end{equation}
Then, they are projected to ray space via a local affine
transformation.
% \begin{equation} \label{preEq.4}
% \bm{\Sigma}_{k}^{''} =\mathbf{J}_{k}\bm{\Sigma}_{k}^{'}\mathbf{J}_{k}^{T},
% \end{equation}
% 
Finally, 3DGS utilizes spherical harmonics to model view-dependent color $\bm{c}_{k}$ and renders images via alpha blending according to the primitive’s depth order:
\begin{equation} \label{preEq.5}
c(x) = \sum^{K}_{k=1}\bm{c_{k}}\alpha_{k}\mathcal{G}^{2D}_{k}\left(x\right)\prod^{k-1}_{j=1}\left(1-\alpha_{j}\mathcal{G}^{2D}_{j}(x)\right),
\end{equation}
where $\mathcal{G}^{2D}$ is the scaled 2D Gaussian, obtained by removing the third row and column of the ray space covariance matrix. In this work, we reconstruct a 3D Gaussian field by inpainting the reference image and then rendering the coarse background video using volume splitting.

\begin{figure}[t]
\vspace{-.4cm}
\setlength{\abovecaptionskip}{-.2cm}
\setlength{\belowcaptionskip}{-.3cm}
\begin{center}
   \includegraphics[width=1.0\linewidth]{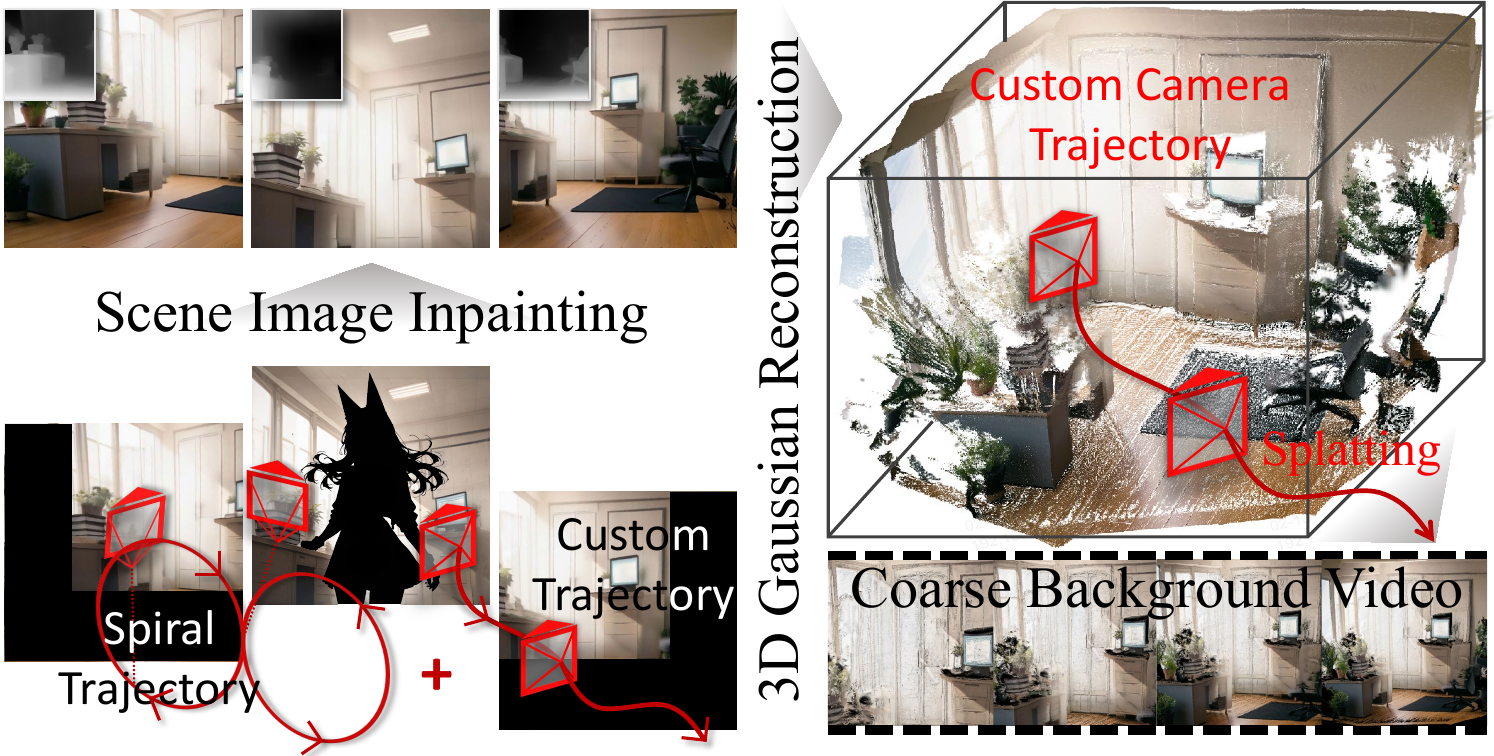}
\end{center}
   \caption{\textbf{Camera-aware Scene Inpainting} for stable scene reconstruction. We use both the pre-defined spiral camera trajectory and the custom camera trajectory to reconstruct a 3D Gaussian field via the warp-and-inpaint scheme.}
\label{fig:3}
\end{figure}

\subsection{Stable Scene Reconstruction}
Existing 3D-agnostic motion guidance makes consistent background generation during large-scale camera movements an ill-posed problem \cite{MikuMikuDance}. Therefore, we reconstruct stable Gaussian scenes to facilitate character art animation.

\vspace{1mm}
\noindent\textbf{Camera-aware Scene Inpainting.} As illustrated in Figure \ref{fig:3}, inspired by the warp-and-inpaint scheme \cite{shriram2024realmdreamer, wang2024vistadream}, we use a pre-defined spiral camera trajectory to reconstruct the 3D scene. Firstly, at the starting point of the camera trajectory, we use LLaVA \cite{liu2024visual} to generate detailed descriptions of the reference background, and use Fooocus \cite{Fooocus} to inpaint the empty regions left by the removed character. Afterward, we estimate the global depth map on this complete background using DepthPro \cite{bochkovskii2024depth}. Next, as the camera moves along the spiral trajectory, we warp the background image to each new viewpoint using its depth map, and then fill the empty regions through image inpainting. After this warp-and-inpaint process, we obtain a set of RGBD images, which are then used to train a stable 3D Gaussian field. The spiral trajectory can be formulated as:
\begin{equation} \label{Eq.1}
\begin{aligned}
\mathbf{P}(t) &= \left[ \begin{array}{c}
     r\cdot sin(2\pi t)\cdot cos(2\pi t) \\
     r\cdot sin(2\pi t)\cdot sin(2\pi t) \\
     - sin(2\pi t)
\end{array} \right], \\
\mathbf{R}(t) &= \left[ \begin{array}{c}
norm(\bm{o}-\bm{p}_{t}) \times \mathcal{U} \\
   \mathcal{U} \\
     norm(\bm{o}-\bm{p}_{t})
\end{array} \right]^{-1} ,
\end{aligned}
\end{equation}
where $\mathbf{P}$ is the position and $\mathbf{R}$ is the rotation of the camera, $t \in [0,1]$ is the camera time step, and $r$ is the radius of the field. $\mathcal{U}$ is the up vector $[0,1,0]$ and $\bm{o}$ is the camera looking point. This well-defined spiral trajectory effectively covers most of the missing regions and generates comprehensive multi-view images. However, the custom camera movements, provided by the user or extracted from the driving video, may differ significantly from the spiral trajectory. Therefore, we expand the 3D Gaussian field according to the custom trajectory using the warp-and-inpaint strategy again. Before this process, the camera rotations are standardized based on the first camera frame to ensure consistency with the spiral trajectory. Finally, based on the custom camera trajectory, a background video is rendered through volume splatting at each camera step.

The reason we do not directly use the custom camera trajectory to reconstruct the 3D scene is that it may be excessively dynamic, potentially resulting in a discontinuous and unstable 3D scene. Additionally, since the reconstructed 3D scene often suffers from fidelity issues, the rendered background video may contain blurring, distortions, and black voids. To address these challenges, we introduce a pose-aware video inpainting strategy in the next step, which not only integrates the animated character but also refines the coarse background video for improved visual quality.

\begin{figure}[t]
\vspace{-.4cm}
\setlength{\abovecaptionskip}{-.2cm}
\setlength{\belowcaptionskip}{-.3cm}
\begin{center}
   \includegraphics[width=1.0\linewidth]{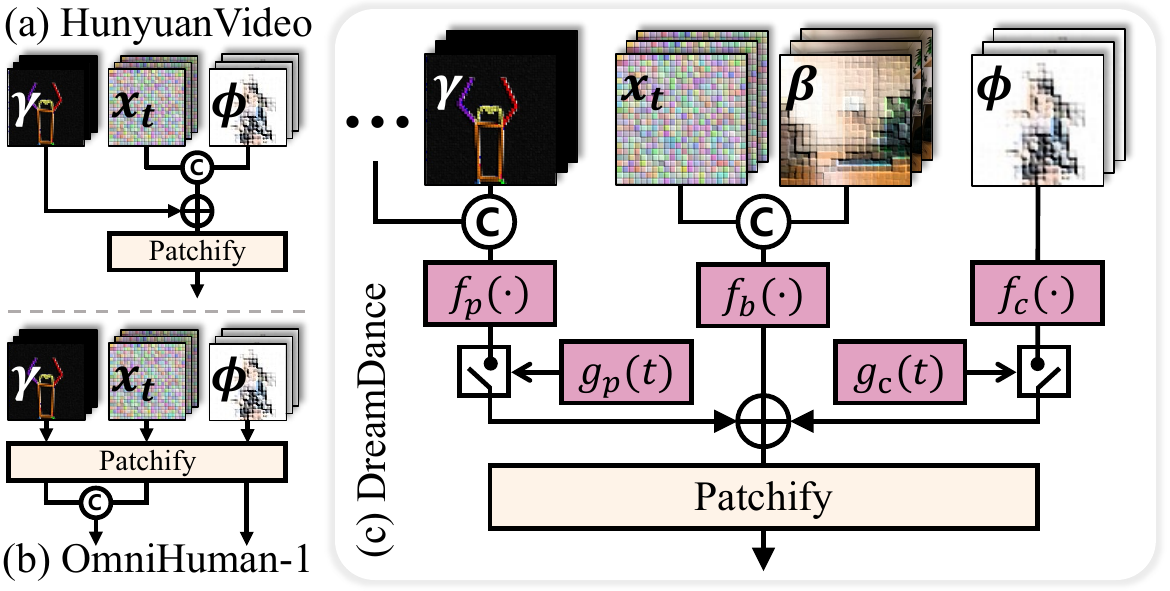}
\end{center}
   \caption{\textbf{The gating strategy} in our MM-DiT model and its comparison with the mainstream condition incorporation methods.}
\label{fig:4}
\end{figure}

\subsection{Dynamic Character Animation}
Based on the coarse background video generated in the first stage, we implement pose-aware video inpainting to achieve dynamic character animation. Previous UNet-based reference-denoising architectures lack the ability to model video coherence effectively \cite{MikuMikuDance}. Therefore, we introduce an MM-DiT-based video foundation model along with a gating strategy to enable pose-guided character integration and ensure temporal consistency.

\vspace{1mm}
\noindent\textbf{Pose-aware Video Inpainting.} As illustrated in Figure \ref{fig:2}, we divide the input references and guidance into three sets. The first is the background set, which includes the coarse background video, and an inpainting mask video generated based on the region of the driving pose. The second is the pose set, consisting of the driving face, hand, and body pose videos. The final set is the reference character. All elements in these three sets are encoded by the pre-trained 3D VAE, and then stacked along the channel dimension to obtain the latent background $\bm{\beta}$, pose $\bm{\gamma}$, and character $\bm{\phi}$. Next, the latent background is concatenated with the base latent noise $\bm{x}_{t}$ along the channel dimension. Then, three convolutional layers are applied to project each of the three latent features to the same channels, respectively.

Existing DiT-based models for character image animation typically use simple feature concatenation or addition to inject the latent reference and guidance \cite{lin2025omnihuman, kong2024hunyuanvideo}. However, unlike these methods, the goal of our model is to inpaint the base background video using the posed character. Obviously, character appearance and pose information should be prioritized during the initial denoising steps, while at later steps, the model should focus more on overall video refinement. To achieve this, we propose two denoising step-based gates that adaptively inject the latent character and pose into the base latent noise according to the denoising step $t$. Each gate consists of a Linear layer followed by a $tanh$ activation function. This gating strategy can be formulated as:
\begin{equation} \label{Eq.2}
\begin{aligned}
\bm{x}_{t}^{'} = &f_{b}\left( \left[\bm{\beta},\bm{x}_{t} \right] \right) \\ &+ tanh\left( g_{p}( t) \right) \cdot f_{p}(\bm{\gamma}) +tanh\left( g_{c}( t) \right) \cdot f_{c}(\bm{\phi}),   
\end{aligned}
\end{equation}
where $f(\cdot)$ denotes the convolutional Project-In layers, and $g(\cdot)$ represents the Linear layers. A detailed comparison of the structural differences between existing methods and our gating strategy is shown in Figure \ref{fig:4}. Finally, we use MM-DiT from CogVideoX \cite{yang2024cogvideox} to perform the diffusion denoising steps. Additionally, the reference character is embedded using the CLIP image encoder \cite{radford2021learning} and serves as the text hidden states in the cross-attention operations of MM-DiT. This process is commonly used in existing work and is therefore omitted from Figure \ref{fig:2}. The resulting latent output is decoded through the 3D VAE Decoder to generate the character art animation.

\vspace{1mm}
\noindent\textbf{3D Gaussian-free Training Approach.}
We perform supervised fine-tuning to train the gated video inpainting model in DreamDance, starting from the image-to-video model CogVideoX-5B. Given that constructing 3D Gaussian fields is time-consuming, we inpaint the character region and apply random down-sampling to the original video background to simulate the coarse video rendered from the 3D Gaussian scene. The down-sampling approach includes adding black blocks, introducing noise, blurring, and applying random perspective transformations. Additionally, following \cite{MikuMikuDance}, we randomly generate stylized pair-wise images by concatenating the initial frames along the spatial dimension and use the depth and edge-controlled SDXL-Neta model \cite{neta} to transfer the art style. Then, the stylized frames are repeated along the temporal dimension to construct a fake video for training. To simulate the inference process, in which the reference character art is irrelevant to the driving pose, we randomly select reference frames that are not involved in the target video clips.

During the training, we found that the supervision of the background region in the video was too strong, causing the bodies of the inpainted characters to be incomplete. To address this issue, we use inpainting masks to re-weight the loss and fill the character bounding boxes of the background videos with black in the early training stages, thereby enhancing dynamic character learning and eliminating the model from overfitting to the backgrounds.
\section{Experiments}

\noindent\textbf{Datasets.}
To train DreamDance, we collected an MMD video dataset comprising 4,800 animations created by artists, which is comparable to that of MikuDance. We split these videos into approximately 150,000 clips, which together include over 14.8 million frames. For the quantitative evaluation, we used 100 MMD videos that were not included in the training set, with their first frames serving as reference images. For the qualitative evaluation, all character art was randomly generated using SDXL-Neta \cite{neta}, and the driving videos were not seen during training.

\vspace{1mm}
\noindent\textbf{Implementation details.}
We implement DreamDance using the code base of VistaDream \cite{wang2024vistadream} and Finetrainers \cite{finetrainers}. Experiments are conducted on 32 NVIDIA A800 GPUs. During training, the videos are center-cropped and resized to a resolution of $768 \times 768$, and the length is sampled to 48 frames. Training is conducted for 60,000 steps with a batch size of 32. The learning rates are set to 1e-4, and the dropout ratio for the character and pose guidance is set to 0.1. During inference, we use a DDIM sampler for 50 denoising steps. We adopt the temporal aggregation method described in \cite{tseng2023edge} to generate long videos. The code will be released in the final version.

\vspace{1mm}
\noindent\textbf{Evaluation metrics.}
Following MikuDance \cite{MikuMikuDance}, we evaluate the results from two aspects: image and video. To assess image quality, we report frame-wise FID \cite{heusel2017gans}, SSIM \cite{wang2004image}, LISPIS \cite{zhang2018unreasonable}, PSNR \cite{hore2010image}, and and L1. For video quality, we concatenate every consecutive 16 frames to form a sample, from which we report FID-VID \cite{balaji2019conditional} and FVD \cite{unterthiner2018towards}.

\subsection{Qualitative Results}

\begin{figure*}[t]
\vspace{-.8cm}
\setlength{\abovecaptionskip}{-.3cm}
\setlength{\belowcaptionskip}{-.2cm}
\begin{center}
   \includegraphics[width=1.0\linewidth]{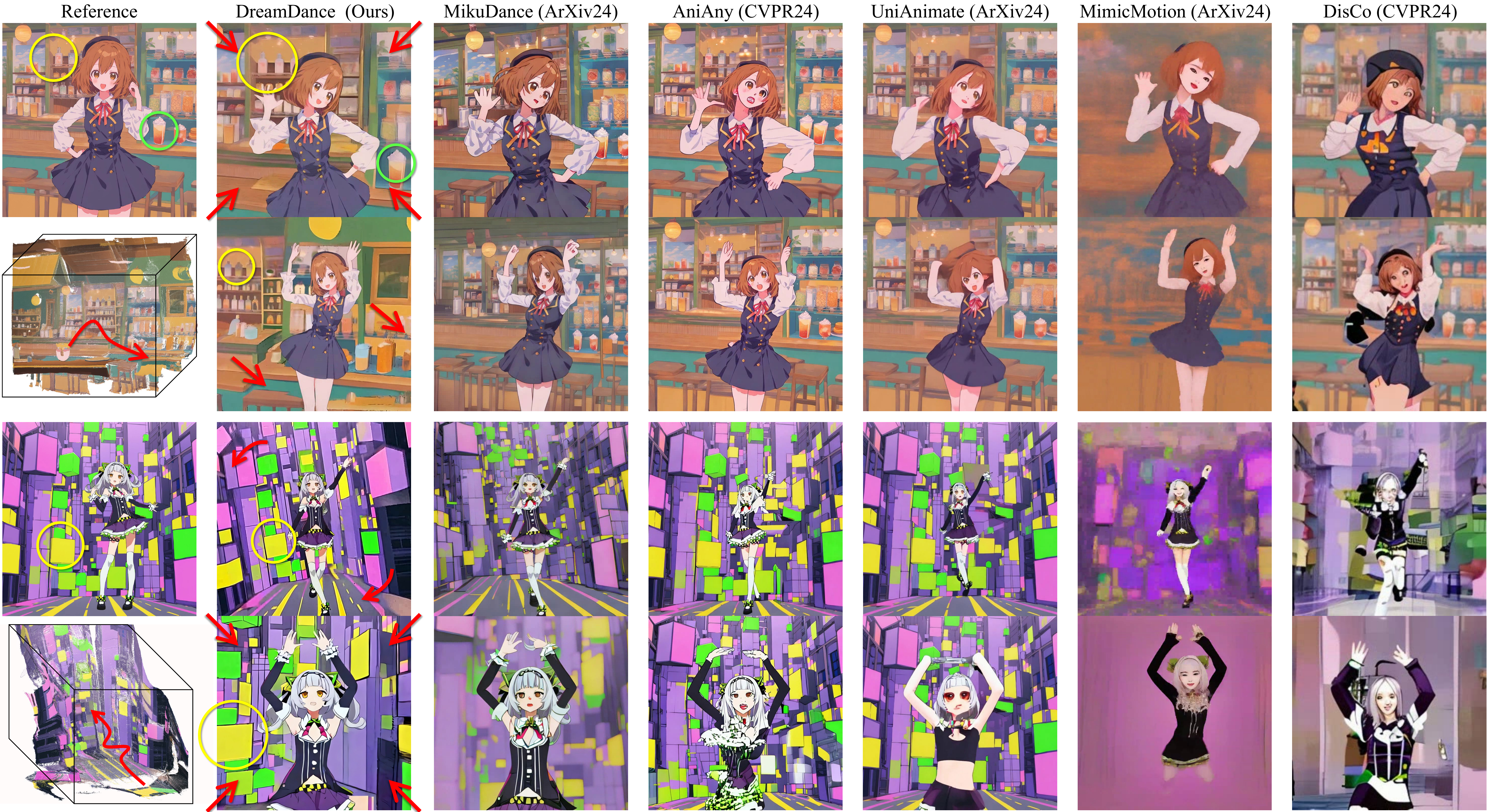}
\end{center}
   \caption{\textbf{Comparison with image animation baselines.} The red arrows represent the approximate direction of camera movement, while the circles highlight significant correspondences. The black boxes contain the reconstructed 3D Gaussian scenes of our DreamDance.}
\label{fig:exp_sota}
% \end{figure*}

\setlength{\abovecaptionskip}{-.3cm}
\setlength{\belowcaptionskip}{-.3cm}
\begin{center}
   \includegraphics[width=1.0\linewidth]{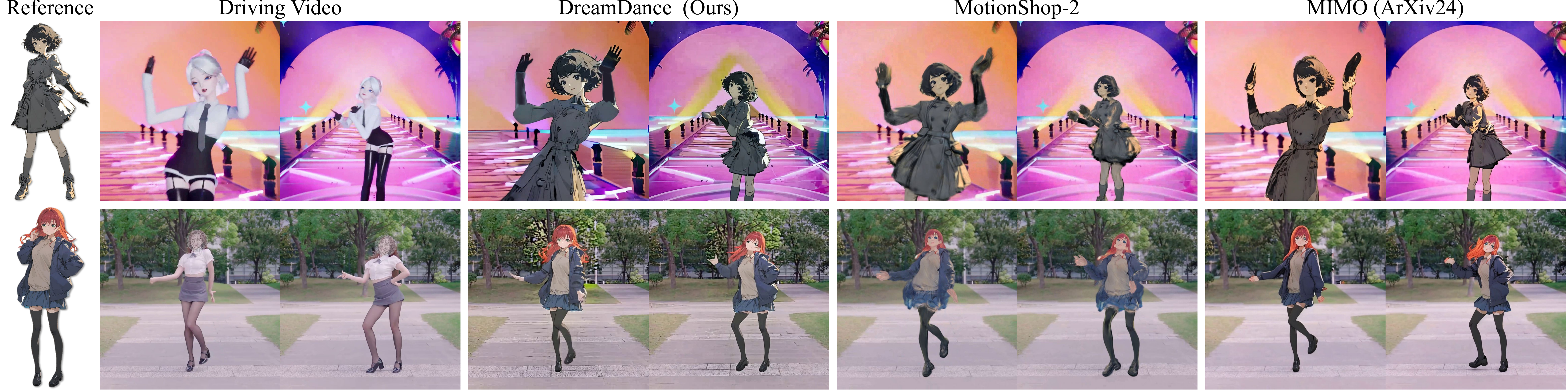}
\end{center}
   \caption{\textbf{Comparison with character replacement baselines.} MIMO and MotionShop-2 only support full-body reference images.}
\label{fig:exp_video}
\end{figure*}

\noindent\textbf{Comparison with image animation baselines,} including MikuDance \cite{MikuMikuDance}, as well as recent human animation methods, such as Animate Anyone (AniAny) \cite{hu2024animate}, UniAnimate \cite{wang2024unianimate}, MimicMotion \cite{zhang2024mimic}, and DisCo \cite{wang2024disco}.

The results in Figure \ref{fig:exp_sota} demonstrate that AniAny, UniAnimate, MimicMotion, and DisCo struggle with a strong shape prior on the human body, which leads to substantial character distortion and fidelity issues in their outputs. Moreover, the backgrounds in their generated videos remain nearly static or excessively blurry, resulting in monotonous and visually flat effects. While DisCo employs an independent ControlNet to process the backgrounds, it suffers from scene collapse when animating character art. MikuDance shows significant improvements in animating character art, but the backgrounds still exhibit inconsistencies when confronted with large-scale camera movements. Notably, thanks to the explicit reconstruction of the 3D scene and the MM-DiT-based inpainting paradigm, our DreamDance achieves precise camera control, coherent scene dynamics, and consistent animation generation, producing high-quality and vivid 2D animation results.

\vspace{1mm}
\noindent\textbf{Comparison with character replacement baselines.}  One valuable application of DreamDance is its ability to directly replace humans in driving videos with reference characters. We compare it with MotionShop-2 \cite{Motionshop-2} and MIMO \cite{men2024mimo}, which are 3D and 2D-based methods, respectively. As shown in Figure \ref{fig:exp_video}, MotionShop-2 exhibits noticeable character distortion due to the unresolved challenges of 3D character reconstruction, while MIMO fails to effectively preserve the attributes of the reference characters. Additionally, both MotionShop-2 and MIMO only support full-body character images. In contrast, our DreamDance seamlessly integrates the 2D character into the driving video without disrupting the harmony of the scene. This application opens up broad prospects for DreamDance in creating flexible video content.

\begin{figure*}[t]
\vspace{-.8cm}
\setlength{\abovecaptionskip}{-.3cm}
\setlength{\belowcaptionskip}{-.2cm}
\begin{center}
   \includegraphics[width=1.0\linewidth]{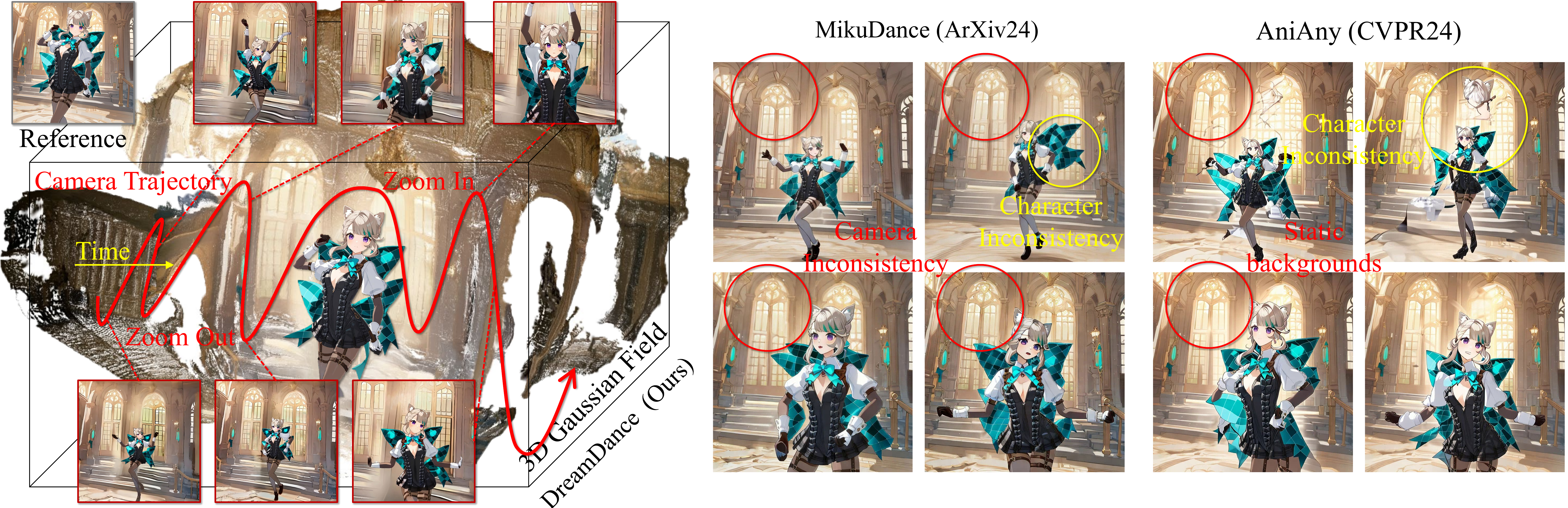}
\end{center}
   \caption{\textbf{High-dynamic and precise camera control} of our DreamDance. MikuDance exhibits inconsistencies due to its 3D-agnostic motion guidance, while AniAny produces static backgrounds. In contrast, DreamDance generates coherent and vivid animations.}
\label{fig:exp_camera}
% \end{figure*}

\setlength{\abovecaptionskip}{-.3cm}
\setlength{\belowcaptionskip}{-.3cm}
\begin{center}
   \includegraphics[width=1.0\linewidth]{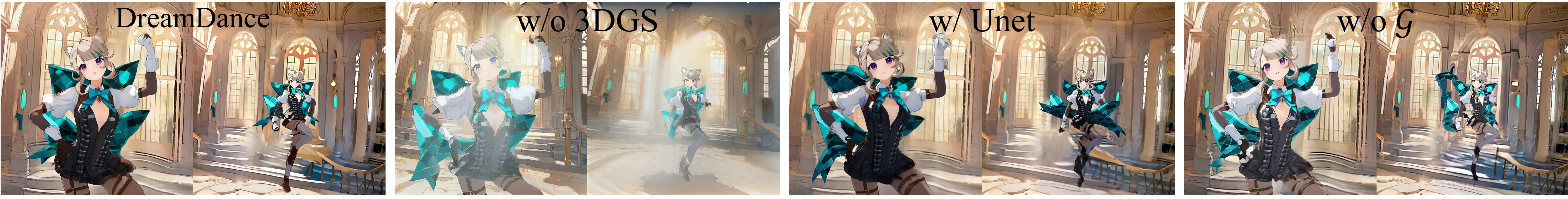}
\end{center}
   \caption{\textbf{Ablation experiments.} ``w/o 3DGS'', ``w/ UNet'', and ``w/o $\mathcal{G}$'' are defined in Section \ref{ablsty}.}
\label{fig:exp_abl}

\setlength{\abovecaptionskip}{-.3cm}
\setlength{\belowcaptionskip}{-.3cm}
\begin{center}
   \includegraphics[width=1.0\linewidth]{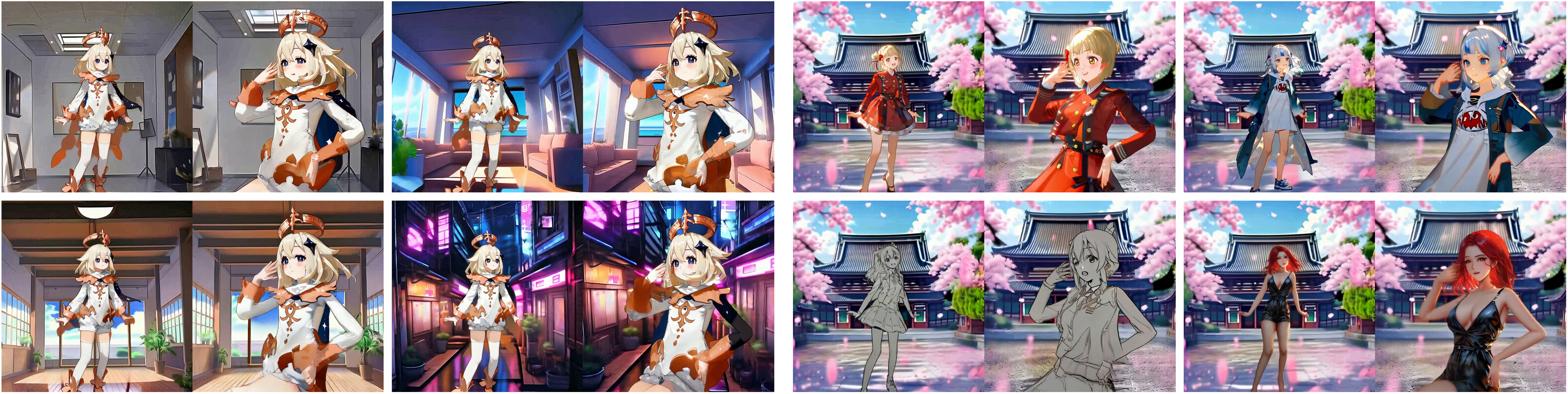}
\end{center}
   \caption{\textbf{Generalizability on various scenes (left) and characters (right).} Please see our demo video for a clearer understanding.}
\label{fig:exp_scene_char}
% \end{figure*}
\end{figure*}

\vspace{1mm}
\noindent\textbf{High-dynamic and precise camera control.} A key highlight of DreamDance is its ability to animate characters with high-dynamic camera movements while maintaining scene coherence through precise camera control. Distinct from MikuDance, which relies on 2D flow for scene motion guidance, and AniAny, which always outputs static backgrounds, DreamDance explicitly reconstructs stable 3D scenes. This approach avoids the context ambiguity that typically arises in the scene inpainting process, ensuring more consistent and immersive character animation with high-dynamic motion, as demonstrated in Figure \ref{fig:exp_camera}.

\begin{figure}[t]
\vspace{-.4cm}
\setlength{\abovecaptionskip}{-.2cm}
\setlength{\belowcaptionskip}{-.5cm}
\begin{center}
   \includegraphics[width=1.0\linewidth]{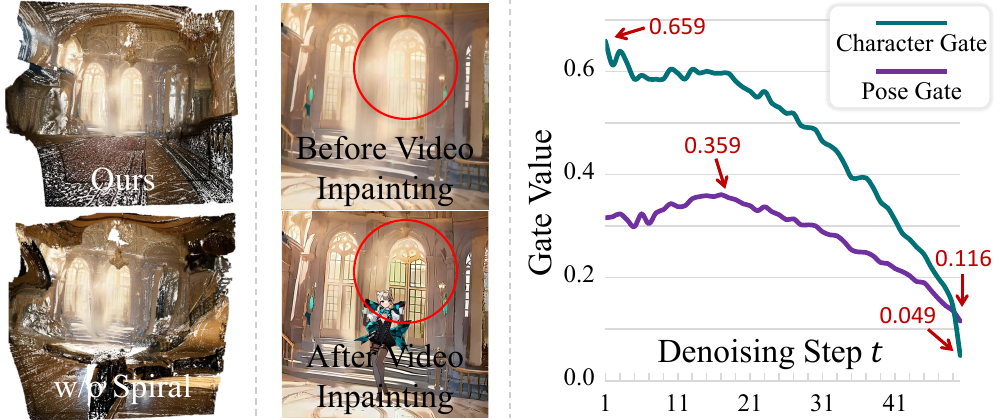}
\end{center}
   \caption{\textbf{Ablations on spiral trajectory (left), Scene enhancement (middle), and visualization of the gate values (right).}}
\label{fig:exp_spiral}
\end{figure}

\vspace{1mm}
\noindent\textbf{Ablation study.} In Figure \ref{fig:exp_abl} and Figure \ref{fig:exp_spiral}, we conduct ablation experiments to verify the key designs of our DreamDance, which include the camera-aware scene inpainting method, the MM-DiT-based model architecture, and the gating strategy for video inpainting. \label{ablsty}

To demonstrate the necessity of the camera-aware scene inpainting in DreamDance, we implemented a baseline (w/o 3DGS) that animates the character art directly using the MM-DiT model, bypassing the 3D Gaussian reconstruction process. The results in Figure \ref{fig:exp_abl} indicate that this model fails to generate consistent backgrounds due to the implicit inpainting of unknown regions, leading to flickering and blurring in the generated video. Moreover, as shown in  Figure \ref{fig:exp_spiral}, reconstructing the 3D scene without the spiral camera trajectory may result in discontinuous Gaussian fields.

To evaluate the MM-DiT-based model architecture, in Figure \ref{fig:exp_abl}, we implemented a UNet-based video inpainting model. Since this model has limited capabilities in spatio-temporal modeling and a smaller pre-training scale compared to MM-DiT, its results are inferior to those of DreamDance. To assess the effectiveness of the gating strategy in our pose-aware video inpainting step, we conducted an experiment in which the latents were directly added to the noise without re-weighting them using the gates (w/o $\mathcal{G}$). This ablation model was trained on the same dataset with the same training settings as DreamDance. However, its results exhibited an underfitting phenomenon, with the videos appearing blurry during high-dynamic motion. Moreover, as shown in Figure \ref{fig:exp_spiral}, our gated video inpainting model also enhances the quality of the backgrounds provided by the rendered 3D scene for character art animation.

We visualize the values of the adaptive gates across the denoising time steps in the right part of Figure \ref{fig:exp_spiral}. As the denoising steps progress from 0 to 50, both the character gate and the pose gate decrease. This supports our conjecture that the model requires more information about the character’s appearance and pose during the early denoising steps, whereas in the later stages, it prioritizes optimizing the quality of the existing latent videos.

\vspace{1mm}
\noindent\textbf{Generalizability on various scenes and characters.} Beyond reconstructing the scene from the reference character art, DreamDance supports scene reconstruction from custom images and the animation of the reference character across various scenes. As shown in the left part of Figure \ref{fig:exp_scene_char}, DreamDance effectively integrates animated characters into diverse 3D Gaussian scenes. On the other hand, DreamDance is also capable of handling multiple characters in a wide range of art styles, including but not limited to celluloid, antiquity, and line sketch, as demonstrated in the right part of Figure \ref{fig:exp_scene_char}. This high level of flexibility opens up vast possibilities for 2D animation applications.

% Please add the following required packages to your document preamble:
% \usepackage{multirow}
\begin{table*}[t]
% \setlength{\abovecaptionskip}{-.01cm}
% \setlength{\belowcaptionskip}{-.2cm}
% \vspace{-.5cm}
\centering
\small{
\caption{ \textbf{Quantitative comparisons with baselines and ablative experiments.}  `UNet' and `w/o $\mathcal{G}$' are defined in Section \ref{ablsty}. `Foreground-only' refers to replacing the reference backgrounds with white images and evaluating only the character animations. The best results are highlighted in bold, and the second-best are underlined. DreamDance achieves superior results across most metrics.}%
\label{tab:1}
\begin{tabular}{cl|ccccc|cc|c|cc}
\multicolumn{2}{c|}{\textbf{Methods}} & \textbf{SSIM}$_{\uparrow}$ & \textbf{PSNR}$_{\uparrow}$ & \textbf{LISPIS}$_{\downarrow}$ & \textbf{L1}$_{\downarrow}^{E-05}$ & \textbf{FID}$_{\downarrow}$  & \textbf{FID-VID}$_{\downarrow}$ & \textbf{FVD}$_{\downarrow}$ & \multirow{9}{*}{\rotatebox{90}{\textbf{Foreground-only}}} & \textbf{FID}$_{\downarrow}$ & \textbf{FVD}$_{\downarrow}$ \\ \cmidrule{1-9} \cmidrule{11-12} 
\multicolumn{1}{c|}{\multirow{2}{*}{\rotatebox{90}{\textbf{SVD}}}} & UniAnimate \cite{wang2024unianimate} & 0.417 & 12.074 & 0.571 & 7.930 & 47.328 & 40.924 & 882.245 & & 29.818 & 381.485 \\
\multicolumn{1}{c|}{} & MimicMotion \cite{zhang2024mimic} & 0.325 & 12.264 & 0.600 & 9.313 & 60.210 & 44.517 & 903.674 &  & 30.125 & 407.856 \\ \cmidrule{1-9} \cmidrule{11-12} 
\multicolumn{1}{c|}{\multirow{4}{*}{\rotatebox{90}{\textbf{SD}}}}  & DisCo \cite{wang2024disco} & 0.313 & 10.732 & 0.615 & 9.248 & 59.221 & 46.852 & 923.921 &  & 31.221 & 564.892 \\
\multicolumn{1}{c|}{} & AniAny \cite{hu2024animate} & 0.488 & 12.530 & 0.548 & 7.307 & 43.945 & 38.179 & 846.414 &  & 27.927 & 326.842 \\
\multicolumn{1}{c|}{} & MikuDance \cite{zhang2024mikudance} & 0.576 & 14.592 & 0.493 & 5.726 & \textbf{24.597} & 22.868 & 502.380 &  & \textbf{14.835} & \underline{194.124} \\
\multicolumn{1}{c|}{} & DreamDance UNet & 0.612 & 16.721 & 0.383 & 4.622 & 32.923 & 19.387 & 477.235 &  & \underline{15.227} & 221.126 \\ \cmidrule{1-9} \cmidrule{11-12} 
\multicolumn{1}{c|}{\multirow{2}{*}{\rotatebox{90}{\textbf{DiT}}}} & DreamDance w/o $\mathcal{G}$ & \underline{0.626} & \underline{17.135} & \underline{0.378} & \underline{4.601} & 30.794 &  \underline{17.198} & \underline{441.057} &  & 16.831 & 217.946 \\
\multicolumn{1}{c|}{} & DreamDance (Ours) & \textbf{0.699} & \textbf{17.964} & \textbf{0.355} & \textbf{4.109} & \underline{29.659} & \textbf{16.411} & \textbf{430.136} &  & 16.102 & \textbf{188.852} \\ \cmidrule{1-9} \cmidrule{11-12}
\end{tabular}}
\end{table*}

\subsection{Quantitative Results}
Table \ref{tab:1} presents quantitative comparisons and the results demonstrate that DreamDance achieves state-of-the-art performance across most image and video metrics. Additionally, the ablation results confirm the effectiveness of the key design elements in the dynamic character animation stage of DreamDance. To isolate character quality from background effects, we conducted evaluations on foreground-only results, as shown in the right part of Table \ref{tab:1}. In this setup, we replaced the backgrounds of reference character art with white images for baseline evaluation. For DreamDance, we provided white background videos for character animation inpainting. Under these conditions, DreamDance consistently achieved the best video temporal quality.

\begin{figure}[t]
\vspace{-.4cm}
\setlength{\abovecaptionskip}{-.2cm}
\setlength{\belowcaptionskip}{-.5cm}
\begin{center}
   \includegraphics[width=1.0\linewidth]{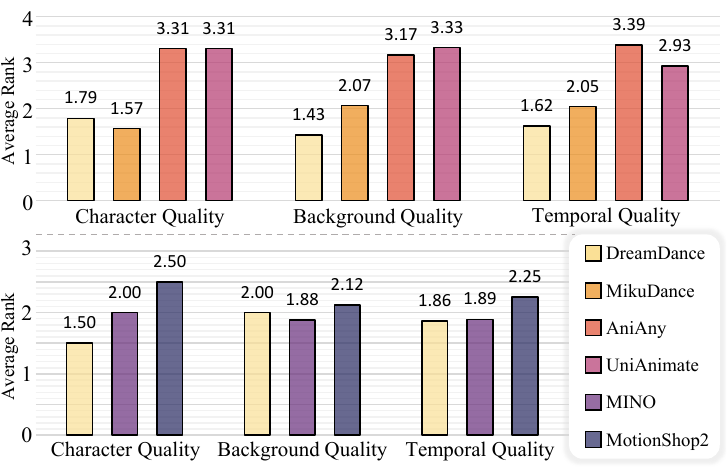}
\end{center}
   \caption{\textbf{User Study} for image animation (top) and character replacement (bottom). The smaller value means the better quality.}
\label{fig:userstudy}
\end{figure}

\vspace{1mm}
\noindent\textbf{User study.} We invited 50 volunteers to evaluate DreamDance against baseline methods on two tasks: image animation and character replacement. Each participant reviewed 15 videos, each containing one pose guidance and three or four anonymous animation results. They ranked the results based on character quality, background quality, and temporal consistency. After filtering out abnormal responses, the average rankings are summarized in Figure \ref{fig:userstudy}. For image animation, DreamDance significantly outperforms baseline methods in background and temporal quality, with over 77.27\% of users preferring its animations. For character replacement, DreamDance achieves the highest character and temporal quality, favored by more than 59.09\% of users.

\section{Conclusions}
In this work, we propose DreamDance, a new inpainting-based pipeline for animating character art. DreamDance integrates two key techniques: Camera-aware Scene Inpainting and Pose-aware Video Inpainting. Camera-aware Scene Inpainting reconstructs stable Gaussian scenes, allowing for the rendering of coarse yet context-coherent background videos. Pose-aware Video Inpainting then adaptively incorporates pose-guided characters into the background, refining the video quality and ensuring consistent animation for stylized character art. Extensive experiments demonstrate that DreamDance outperforms baseline methods, achieving state-of-the-art results in character art animation.

\noindent\textbf{Limitations.} We acknowledge that some generated animations exhibit artifacts, particularly in character details such as the hands and clothing. This issue arises from the limitations of the datasets and the base model. Additionally, extracting precise camera parameters from real-world videos remains a challenge, often requiring manual adjustments.

{
    \small
    \bibliographystyle{ieeenat_fullname}
    \bibliography{main}
}

\end{document}